\documentclass[10pt,twocolumn,letterpaper]{article}

\usepackage{cvpr}              
\usepackage{graphicx}

\usepackage{booktabs}
\usepackage{multirow}
\usepackage{makecell}

\definecolor{cvprblue}{rgb}{0.21,0.49,0.74}
\usepackage[pagebackref,breaklinks,colorlinks,allcolors=cvprblue]{hyperref}

\def\confName{CVPR}
\def\confYear{2027}

\title{Beyond Binary Correctness: A Surrogate for Confidence Learning in \\ Large Language and Vision-Language Models}

\author{
Chen Li\textsuperscript{1}\thanks{Email: Chen Li (li.chen.8@stonybrook.edu).}\hskip 1em
Xiaoling Hu\textsuperscript{2}\hskip 1em
Chao Chen\textsuperscript{1}\hskip 1em
\\
\textsuperscript{1}Stony Brook University, NY, USA \hskip 1em
\\
\textsuperscript{2}Massachusetts General Hospital and Harvard Medical School, MA, USA \hskip 1em
}

\begin{document}

\newcommand{\myparagraph}[1]{\smallskip\noindent\textbf{#1}}


\maketitle
\begin{abstract}
Confidence estimation for large language models (LLMs) and
vision-language models (VLMs) aims to assess how likely a particular
generated response is to be correct. We investigate how learning
this assessment can benefit from the model's predictive reliability
on the underlying query.
Binary correctness supervision, however, assigns identical labels to occasional errors and frequent failures, as well as isolated successes and consistent success. These outcomes can arise from substantially different predictive behavior under stochastic decoding. We propose a confidence-learning framework that supplements realized response correctness with query-level predictive reliability. We estimate reliability through repeated generations under a fixed model and decoding configuration, then combine the empirical success rate with individual correctness to construct a graded confidence surrogate. The surrogate differentiates responses sharing the same correctness label while preserving the average predictive reliability of the sampled response pool. To learn this target, we condition an independent confidence estimator on the query and its full generated response, and optimize backbone LoRA adapters together with a projection and scalar confidence head using mean squared error. Repeated sampling is required only for supervision construction; inference scores an individual response without additional generations. We establish ordering and mean-preservation properties of the surrogate and derive a selective-risk bound connecting population surrogate prediction error to the error rate among accepted responses. We evaluate the framework across language and vision-language models, examining confidence calibration and selective prediction.
\end{abstract}    
\section{Introduction}

Large language models (LLMs) and vision-language models (VLMs) have achieved strong performance across a wide range of tasks~\cite{qwen3technicalreport, Qwen3-VL}, yet they continue to produce incorrect answers and flawed reasoning~\cite{ji2023survey, guan2024hallusionbench}. Average task accuracy alone does not tell users whether a particular generated response should be trusted. Confidence estimation addresses this need by providing scores that help users and downstream systems decide which responses to accept, review, or reject~\cite{srinivasan2024selective, kamath2020selective}. The usefulness of these decisions depends on whether confidence meaningfully reflects the quality of the response being assessed~\cite{guo2017calibration}. Developing reliable confidence estimators is therefore an important step toward dependable use of generative models.

Confidence can be communicated through natural language expressions, generated numerical values, or scalar scores predicted directly from model representations~\cite{lin2022teaching, azaria2023internal}. These formats differ in how the score is produced, but share a common objective: providing an informative assessment of a particular response given its query. Verbalized confidence is attractive because it can be elicited through ordinary language prompts~\cite{tian2023just}, while direct scalar prediction provides an explicit numerical readout. Learning useful confidence also requires informative supervision~\cite{li2026conftuner}, raising a fundamental question: what should the training target convey about the response being assessed?

Existing confidence-learning methods address this question through different training objectives. ConfTuner optimizes a tokenized Brier score against binary answer correctness~\cite{li2026conftuner}, while ADVICE constructs correct and incorrect answer pairs and imposes answer-dependent constraints on their confidence distributions~\cite{seo2026advice}. LACIE uses preferences that account for both answer correctness and listener acceptance~\cite{stengel2024lacie}. These approaches advance confidence learning through scoring objectives, answer conditioning, and confidence communication. We investigate an additional supervisory signal: the generator's probability of answering an individual query correctly across repeated generations. This quantity captures differences that individual correctness labels do not explicitly encode, including whether an observed success is consistent or exceptional.

This distinction is significant due to the stochastic decoding setting recommended for modern models including Qwen3, Qwen3-VL, and DeepSeek-R1~\cite{qwen3technicalreport,Qwen3-VL,deepseekai2025deepseekr1incentivizingreasoningcapability}. Under a fixed model and decoding configuration, the same query can produce both correct and incorrect responses. Consider two queries with error probabilities of $10\%$ and $70\%$, respectively. Either can yield an incorrect response, which receives the same negative correctness label despite representing an occasional failure in the first case and a frequent failure in the second. Similarly, an isolated success and a consistent success both receive the positive label. We distinguish \emph{realized correctness}, which records the outcome of one response, from \emph{predictive reliability}, which describes the probability that the generator answers the query correctly. Binary correctness remains valid supervision for learning correctness probabilities, but a single observed outcome does not directly reveal this query-level reliability.

\begin{figure}[t]
    \centering
    \includegraphics[width=\linewidth]
        {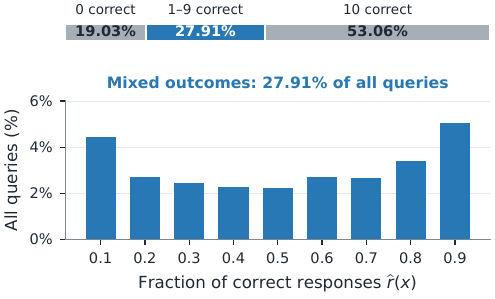}
    \caption{
Correctness variation on M3CoT across $10$ sampled responses
per query from Qwen3-VL-8B-Instruct.
Blue bars indicate mixed correct and incorrect outcomes,
observed in $27.91\%$ of $7{,}861$ training queries.
    }
    \label{fig:m3cot_reliability}
\end{figure}

We examine this variability empirically using $10$ responses sampled from Qwen3-VL-8B-Instruct using its recommended decoding settings for each of $7{,}861$ retained M3CoT training queries. As shown in Figure~\ref{fig:m3cot_reliability}, $2{,}194$ queries ($27.91\%$) yield both correct and incorrect responses. For these queries, the outcome observed in a single generation depends on which response is sampled, and one success or failure does not establish how consistently the generator succeeds. Moreover, the distribution spans different empirical success rates, while binary supervision assigns identical targets to responses with the same realized correctness. These observations motivate supplementing individual correctness with repeated-generation evidence that distinguishes isolated successes from consistent success and occasional errors from frequent failures.


We propose a confidence surrogate that combines answer correctness with
predictive reliability, defined as the model's underlying probability of
answering a question correctly. A single correct or incorrect answer
reveals one outcome, but does not reveal the model's underlying success
probability for that question. During training, we estimate predictive
reliability by sampling multiple answers to each question and computing
the fraction that are correct. We give equal weight to this estimate
and individual answer correctness to construct a graded confidence target.
The added reliability information distinguishes occasional errors from
frequent failures and isolated successes from consistent success.


An existing answer-generation system may need reliable confidence estimates even when its generating model cannot be retrained, or its behavior must remain unchanged. We therefore learn confidence in a separate estimator. This allows confidence-specific supervision to improve answer assessment without modifying the model responsible for producing answers. Because the estimator uses only the original query and generated response, it does not require access to the generator's parameters, hidden states, or token probabilities. Users can consequently add trained confidence estimation to an existing generation pipeline and choose the confidence-estimator backbone independently.

We implement this supervision with a response-conditioned confidence
estimator that directly predicts a scalar score in $[0,1]$. The estimator
receives the original query and the full generated response, including
reasoning when present; the original image is retained for vision-language
tasks. Conditioning on the particular generated answer matches the
answer-dependent nature of our surrogate, allowing different answers to
the same question to receive different confidence scores. Retaining the
reasoning and original input provides the estimator with the evidence
available for assessing that answer. We append an explicit confidence
question and begin the assistant turn with the prefix
\texttt{Confidence:}. The backbone representation at the final token
of this prefix is passed through a learnable projection and scalar output
layer, followed by a sigmoid. Although the input uses a conversational
confidence prompt, the score is produced directly by the scalar head.
This design matches the graded numerical supervision: we optimize the
confidence value directly, without requiring it to be encoded, generated,
and parsed as a textual expression. We train the estimator using mean
squared error against the proposed surrogate. Repeated generations and
reference answers are used to construct training supervision; inference
requires only the query and one generated response.

Our theoretical analysis characterizes how predictive reliability refines the targets. Within the chosen mixing range, the surrogate preserves the target ordering between correct and incorrect responses while introducing reliability-dependent variation within each correctness class. Averaging the surrogate targets over a sampled response pool also recovers that pool's empirical predictive reliability. Furthermore, we show that accurate prediction of the proposed surrogate limits high-confidence mistakes, and derive a theoretical bound on the
fraction of high-confidence answers that are incorrect. These properties explain the construction and its connection to selective prediction. Our experimental study evaluates the practical contribution of predictive reliability through comparisons with existing confidence estimators and controlled ablations against binary correctness supervision, assessing confidence discrimination, selective prediction, and calibration. 

\section{Related Work}
We review confidence estimation for large language models (LLMs) and
vision-language models (VLMs), focusing on confidence elicitation and
sampling, supervision for confidence learning, and confidence prediction
from model representations.

\myparagraph{Confidence elicitation and sampling.}
Confidence estimation for generative models has been studied through verbalized expressions, token probabilities, and evidence from repeated generations~\cite{zhao2024fact, yang2024verbalized, zhou2026steerconf}. Lin et al.~\cite{lin2022teaching} demonstrate that language models can learn to express uncertainty through numerical percentages and linguistic confidence statements. Tian et al.~\cite{tian2023just} investigate prompting strategies for eliciting numerical confidence from models trained with human feedback, showing that generated confidence scores can be better calibrated than answer probabilities derived from the language-model distribution. Xiong et al.~\cite{xiong2024can} systematically evaluate prompting, sampling, and aggregation strategies, identifying both the potential and limitations of verbalized confidence. Sampling-based approaches also exploit agreement and variation among generated answers. Kuhn et al.~\cite{kuhn2023semantic} introduce semantic entropy to measure uncertainty over meanings rather than individual word sequences, accounting for semantically equivalent answers. In vision-language models, Xuan et al.~\cite{xuan2025seeing} examine verbalized calibration across tasks and modalities and introduce visual confidence-aware prompting. These studies investigate how confidence can be elicited or estimated from model outputs. Our work uses correctness across repeated generations to construct training supervision for a confidence estimator.

\myparagraph{Learning confidence through supervision.}
Training-based approaches improve confidence estimation through different supervisory signals and optimization objectives~\cite{bani2026rewarding, damani2026beyond}. ConfTuner~\cite{li2026conftuner} introduces a tokenized Brier score that trains verbalized confidence against binary answer correctness and provides a proper-scoring-rule interpretation of its objective. ADVICE~\cite{seo2026advice} investigates the limited dependence of verbalized confidence on the generated answer and introduces objectives that separate confidence distributions for correct and incorrect answers. LACIE~\cite{stengel2024lacie} formulates confidence communication as a preference-optimization problem, incorporating both answer correctness and acceptance by a simulated listener. SaySelf~\cite{xu2024sayself} constructs supervised confidence targets from the frequency of a reference-verified correct-response cluster, generates self-reflective rationales, and subsequently applies a correctness-based reinforcement-learning reward. Our method combines individual answer correctness and the empirical probability of answering its question correctly within a single graded target. The resulting supervision incorporates predictive reliability for both correct and incorrect answers, distinguishing occasional errors from frequent failures and isolated successes from consistent success.

\myparagraph{Confidence prediction from model representations.}
Model representations provide another means of estimating confidence
without generating a confidence expression as text~\cite{abnar2020quantifying, geva2023dissecting, sundararajan2017axiomatic}.
Azaria et al.~\cite{azaria2023internal} train classifiers on
hidden-layer activations to estimate the truthfulness of statements,
including statements generated by the model itself.
Kadavath et al.~\cite{kadavath2022language} use an additional value
head to predict whether the model can answer a question correctly.
Semantic Entropy Probes~\cite{kossen2024semantic} learn to approximate
semantic entropy from the hidden states of a single generation,
avoiding the repeated sampling required to compute semantic entropy
directly at inference. These approaches demonstrate the use of
internal representations for predicting correctness or uncertainty.
We adopt a scalar confidence estimator conditioned on the question
and the full generated answer, retaining the image for vision-language
tasks. A projection and scalar output layer followed by a sigmoid
produce the numerical confidence score. 

\section{Method}

\begin{figure*}[t]
    \centering
    \includegraphics[width=\textwidth]
        {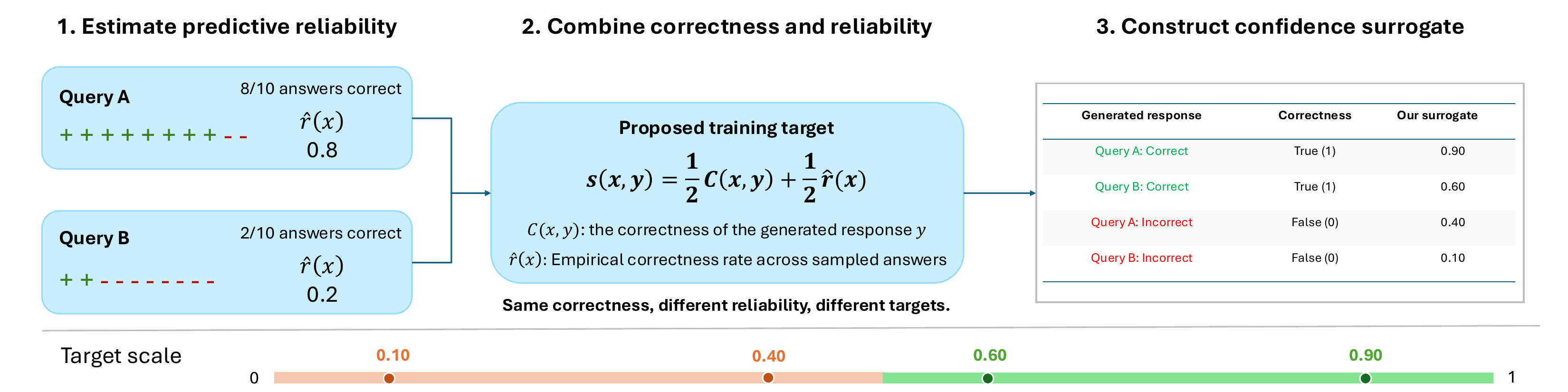}
    \caption{Illustration of the proposed surrogate, combining
    response correctness with empirical predictive reliability.}
    \label{fig:surrogate}
\end{figure*}

\subsection{Motivation}

Non-greedy sampling during inference is explicitly recommended for modern
language and vision-language models, including Qwen3, Qwen3-VL, and
DeepSeek-R1~\cite{qwen3technicalreport,Qwen3-VL,deepseekai2025deepseekr1incentivizingreasoningcapability}.
These configurations retain controlled randomness to support generation
quality. Greedy decoding selects the highest-probability token at each
step, but maximizing local token likelihood does not necessarily produce
the highest-quality complete response. Sampling retains alternative
plausible continuations, while temperature scaling and distribution
truncation regulate their selection. Recent studies demonstrate that
carefully designed sampling strategies improve generation quality and
diversity~\cite{nguyen2025turning,li2026sample}.

We distinguish the correctness of a particular sampled response from
the model's probability of answering the same query correctly across
possible generations. Under stochastic decoding, a fixed model can
produce both correct and incorrect answers to the same query.
An individual correctness label records which outcome occurred in one
generation, but does not directly reveal the model's underlying
probability of answering the query correctly under stochastic decoding.
Consequently, when correctness alone supplies the instance-level
supervision target, distinct patterns of predictive reliability can
receive identical targets.

The variability illustrated in Figure~\ref{fig:m3cot_reliability}
motivates a confidence surrogate that supplements individual correctness
with evidence from the model's behavior across stochastic generations.
The objective is to preserve distinctions that a single binary target
does not explicitly encode: occasional errors versus frequent failures,
and isolated successes versus consistent success. Such a surrogate
provides graded supervision about the reliability of the
response-generating process, enabling confidence learning to account
for both the observed outcome and the broader predictive behavior
underlying it.

\subsection{Proposed Surrogate}

To characterize the model's reliability beyond a single observed outcome, we estimate its probability of success across stochastic generations. For each query $x$, we independently sample $M$ responses $\{y^{(m)}\}_{m=1}^{M}$ while keeping the model and decoding configuration fixed, and compute
\begin{equation}
\widehat{r}(x)
=
\frac{1}{M}\sum_{m=1}^{M}C(x,y^{(m)}),
\label{eq:method_predictive_reliability}
\end{equation}
where $C(x,y^{(m)})\in\{0,1\}$ indicates whether the sampled response is correct according to the reference answer. This empirical rate estimates the model's \emph{predictive reliability} on $x$: its probability of producing a correct answer under the specified stochastic decoding configuration. It provides a graded characterization of query-level reliability, distinguishing queries on which the generator succeeds consistently from those on which success is exceptional.

Predictive reliability alone, however, does not determine the correctness of a particular response. For a fixed query, $\widehat{r}(x)$ is shared by all generated responses, even when some are correct and others are incorrect. A high success probability does not preclude an erroneous response, just as a low success probability does not rule out a valid one. We therefore also consider \emph{realized correctness}, $C(x,y)$, which records whether the particular response $y$ is correct. Predictive reliability characterizes the response-generating process, while realized correctness identifies the outcome of an individual generation. 

Our surrogate should combine these complementary signals to account for both the model's underlying tendency to succeed and the response it actually produces. We define
\begin{equation}
s_{\alpha}(x,y)
=
\alpha C(x,y)
+
(1-\alpha)\widehat{r}(x),
\label{eq:method_confidence_surrogate}
\end{equation}
where $\alpha\in[0,1]$ controls the balance between realized correctness and predictive reliability.

The two terms serve complementary roles. For responses to the same query, the shared reliability term remains fixed, while realized correctness separates correct and incorrect responses by a target difference of $\alpha$. Across queries, the reliability term differentiates responses with the same correctness label according to the generator's probability of success. Thus, an occasional error receives different supervision from a frequent failure, while an isolated success is distinguished from consistent success. The resulting surrogate retains response-specific discrimination while providing graded supervision informed by the broader generation process.


We choose $\alpha\in[\frac{1}{2},1)$ so that correct responses always receive confidence targets greater than or equal to those of incorrect responses, while retaining contributions from both realized correctness and predictive reliability. Since $\widehat{r}(x)\in[0,1]$, correct and incorrect responses satisfy $s_{\alpha}(x,y^{+})\geq\alpha$ and $s_{\alpha}(x',y^{-})\leq1-\alpha$, respectively. Consequently,
\[
s_{\alpha}(x,y^{+})-s_{\alpha}(x',y^{-})
\geq 2\alpha-1\geq0.
\]
Thus, predictive reliability refines the confidence targets within each correctness class without reversing the ordering between correct and incorrect responses. The parameter $\alpha$ controls this trade-off: larger values provide a stronger guaranteed separation between confidence targets, while smaller values within the admissible range retain more variation from query-level reliability. We set $\alpha=\frac{1}{2}$ to give equal weight to realized correctness and predictive reliability.
Figure~\ref{fig:surrogate} illustrates how the surrogate assigns different targets to responses with identical correctness labels by incorporating query-level predictive reliability.


\myparagraph{Preserving predictive reliability.}
The surrogate refines individual targets while preserving the
sampled pool's empirical predictive reliability:
\begin{equation}
\frac{1}{M}\sum_{m=1}^{M}s_{\alpha}(x,y^{(m)})
=\widehat{r}(x).
\label{eq:method_reliability_preservation}
\end{equation}
This identity concerns the constructed targets, rather than
imposing an averaging constraint on learned predictions.
Further derivations and the population-target interpretation
are provided in Supplementary Section A.

\begin{figure*}[t]
    \centering
    \includegraphics[page=1,width=\textwidth]
        {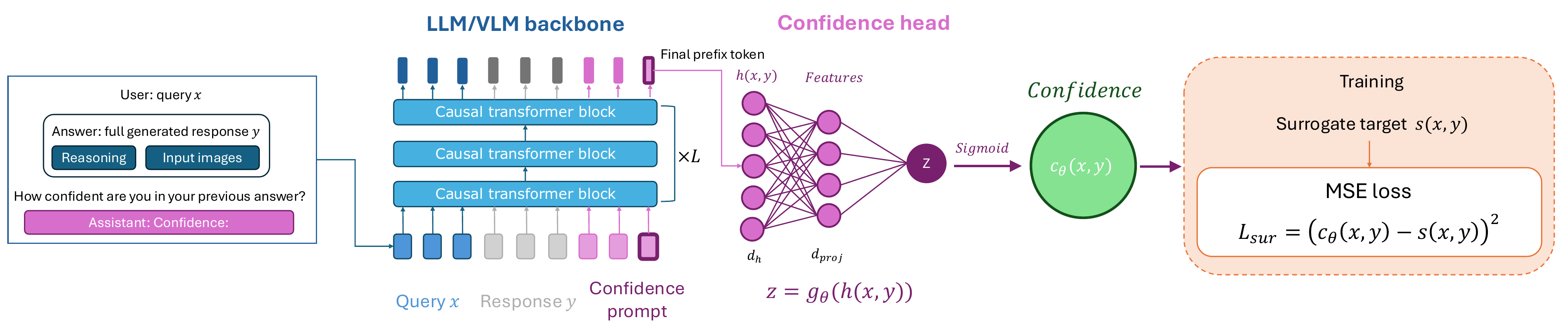}
\caption{
Response-conditioned confidence estimation.
The representation at the final \texttt{Confidence:} prefix token
is mapped to a scalar score through a projection, scalar output layer,
and sigmoid. Training minimizes MSE against the proposed surrogate;
inference requires only the query and one generated response.
}
    \label{fig:confidence_estimator}
\end{figure*}

\subsection{Confidence Estimator}
\label{sec:method_confidence_learning}


Our surrogate depends on both the query and the particular response
generated for it. We therefore construct the confidence estimator's input
from the query $x$ and the full generated response $y$, including its
reasoning when present. For vision-language tasks, $x$ also includes the
original image. For a fixed query, predictive reliability is shared across
responses, but the correctness component of our surrogate depends on the
particular generated answer. A query-only estimator cannot distinguish
these responses, whereas conditioning on $y$ allows confidence to vary
with the observed response. Retaining the complete response makes both
intermediate reasoning and the final answer available as evidence for
confidence assessment. Including the original image similarly allows
the estimator to assess the answer in its visual context.


We append the generated response to the query, followed by the question
\texttt{How confident are you in your previous answer?}, and begin the
assistant turn with \texttt{Confidence:}
(Figure~\ref{fig:confidence_estimator}).
The question directs the backbone to assess the supplied answer,
while the prefix provides a consistent confidence readout location.
The reference answer, correctness, and surrogate are excluded from the input and used only to construct supervision.

We use the final token of the \texttt{Confidence:} prefix as the confidence readout position, whose hidden representations have access to both the query
and the generated response. Placing the readout after the complete answer
ensures that confidence prediction can use all its preceding tokens.
Let $\mathbf{h}(x,y)$ denote the resulting representation supplied to the
confidence head. A learnable projection head maps this contextual
representation to features for confidence regression, and a scalar output
layer produces a score that is mapped to $[0,1]$ through a sigmoid:
\[
c_{\theta}(x,y)
=
\sigma\!\left(g_{\theta}\!\left(\mathbf{h}(x,y)\right)\right),
\]
where $g_{\theta}$ denotes the projection and scalar readout, and
$\sigma$ is the sigmoid function.

Direct scalar prediction matches the numerical form of the proposed
surrogate. It allows the regression objective to penalize the numerical
difference between the predicted confidence and its target directly,
without specifying a vocabulary of confidence expressions or parsing
a generated number. The sigmoid keeps every prediction within the
surrogate's range. The conversational prompt establishes the assessment
context, while the scalar head supplies the numerical prediction used
for training and evaluation. At inference, scoring requires only the
query and one generated response; repeated generations are used to
construct training supervision. Figure~\ref{fig:confidence_estimator} illustrates the response-conditioned estimator, which predicts confidence
from the original query and generated response and is
trained against our surrogate targets.

\myparagraph{Training objective.}
We train a confidence estimator $c_{\theta}(x,y)\in[0,1]$ to predict the surrogate associated with a query and its generated response. For $N$ training examples, the objective is
\begin{equation}
\mathcal{L}_{\mathrm{sur}}(\theta)
=\frac{1}{N}\sum_{i=1}^{N}
\bigl[c_{\theta}(x_i,y_i)-s_{\alpha}(x_i,y_i)\bigr]^2.
\label{eq:method_surrogate_mse}
\end{equation}

\myparagraph{Theoretical analysis.}
Accurate prediction of the proposed surrogate restricts high-confidence
mistakes because every such mistake necessarily creates a minimum
discrepancy between the predicted score and the surrogate target.
We quantify this restriction by deriving an upper limit on the fraction
of high-confidence answers that are incorrect.
The formal selective-risk guarantee and its proof are provided
in Supplementary Section B.

\section{Experiments}
\label{sec:experiments}

We evaluate whether the proposed supervision improves confidence estimation
for both vision-language models (VLMs) and language models (LLMs), and whether
these improvements transfer beyond the dataset used for training. We further
examine the contributions of the two supervisory signals and the design of
the confidence estimator.

\subsection{Experimental Setup}
\label{sec:experimental_setup}
More experimental details are in Supplementary Section C.

\myparagraph{Models and datasets.}
For VLM experiments, we use Qwen3-VL-8B-Instruct as the foundation model.
Confidence learning is performed on M3CoT~\cite{chen2024m3cot}, followed by evaluation on M3CoT,
ChartQA~\cite{masry2022chartqa}, and MMMU~\cite{yue2024mmmu}. For LLM experiments, we consider Qwen3-8B and
Llama-3.1-8B-Instruct. Each model is trained for confidence estimation on
HotpotQA~\cite{yang2018hotpotqa} and evaluated on HotpotQA, GSM8K~\cite{cobbe2021training}, StrategyQA~\cite{geva2021did}, and TriviaQA~\cite{joshi2017triviaqa}.
Thus, M3CoT and HotpotQA provide the source dataset evaluations for the two
modalities, respectively, while the remaining benchmarks assess transfer
without additional confidence training on the target datasets.

\myparagraph{Comparison methods.}
We compare our method with the Base confidence baseline,
Self-Consistency~\cite{xiong2024can},
SaySelf~\cite{xu2024sayself}, LACIE~\cite{stengel2024lacie},
ConfTuner~\cite{li2026conftuner}, and ADVICE~\cite{seo2026advice}.
\textbf{Base} denotes the original foundation model prompted to verbalize a numerical confidence score for its generated response, without confidence-specific training.
These comparisons cover confidence estimation without confidence-specific
training, inference-time repeated sampling, and training-based approaches
using self-reflection, preference optimization, scoring objectives, or
answer-dependent supervision. Trainable methods use M3CoT in the VLM setting
and HotpotQA in the LLM setting. 

\myparagraph{Confidence learning and scoring.}
Our estimator is trained to regress the surrogate combining realized
correctness and empirical predictive reliability. We use equal weights
($\alpha=0.5$) for the main results and examine alternative weights in the
ablation study. The scoring input consists of the original query and its
full generated response, including reasoning when present; the image is
retained for VLM tasks. Repeated generations and reference answers construct
the training targets. At inference, confidence scoring requires only the
query and one generated response. Our estimator scores the base generator's
responses, so its answer accuracy is identical to that of Base.

\myparagraph{Evaluation metrics.}
We assess confidence against binary response correctness using expected
calibration error~\cite{naeini2015obtaining} (ECE $\downarrow$), negative log-likelihood~\cite{guo2017calibration} (NLL $\downarrow$), and the Brier score~\cite{glenn1950verification}.
We measure discrimination between correct and incorrect responses using
AUROC~\cite{hendrycks2017baseline}. For selective prediction, responses are ordered by decreasing
confidence, and we report the area under the risk--coverage curve (AURC $\downarrow$)
and its excess above the oracle ordering~\cite{geifman2019bias} (EAURC $\downarrow$). Lower values are better
for all metrics except AUROC $\uparrow$ and answer accuracy (ACC $\uparrow$). In this paper, $\downarrow$ and $\uparrow$ indicate that lower and higher values are better, respectively. The tables display
ECE multiplied by $100$, and AURC and EAURC multiplied by $10$; other metrics
are unscaled. 

\subsection{Main Results}
\label{sec:main_results}



\myparagraph{Results on vision-language tasks.}
Table~\ref{tab:vlm} demonstrates that our method provides more reliable
confidence estimates for VLM-generated answers. It outperforms all
baselines across every confidence metric on M3CoT and ChartQA, and
achieves the best performance on five of six confidence metrics on
MMMU. Compared with ConfTuner, the strongest discrimination baseline,
our method improves AUROC by 12.0, 5.3, and 3.6 percentage points on
M3CoT, ChartQA, and MMMU, respectively. These gains accompany lower
NLL, Brier score, AURC, and EAURC, showing that our advantage extends
across confidence calibration, error discrimination, and selective
prediction.

Crucially, these improvements require neither modifying the
answer-generating model nor regenerating its responses: our independent
estimator improves confidence assessment while preserving the original
answers. Moreover, training only on M3CoT yields strong results on
ChartQA and MMMU, demonstrating that the learned confidence assessment
transfers beyond the training dataset. Together, these results establish
the practical advantage of our approach: more reliable assessment of
existing model outputs across multiple vision-language tasks.

\begin{figure*}[t]
    \centering
    \includegraphics[width=\textwidth]
    {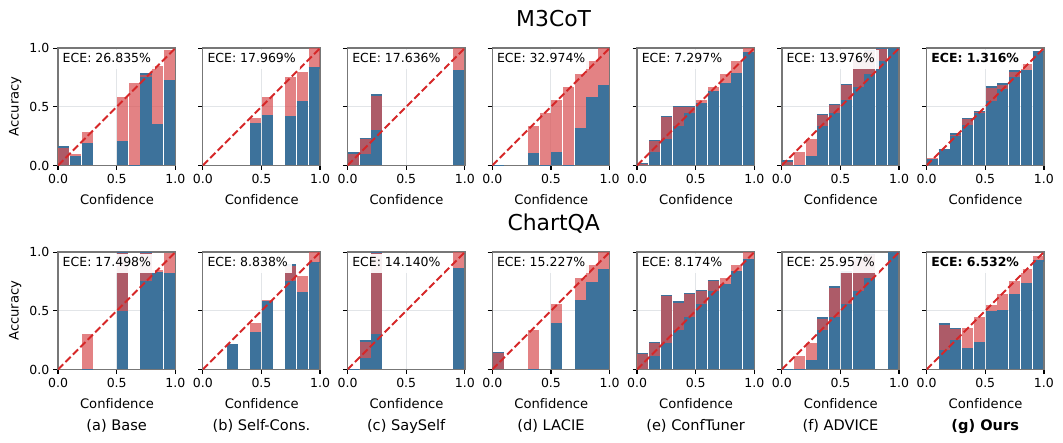}
    \caption{
    Reliability diagrams on M3CoT and ChartQA.
    Blue bars show observed accuracy within ten equal-width
    confidence bins, and red regions indicate the gap between
    accuracy and mean predicted confidence.
    The dashed diagonal denotes perfect calibration.
    Empty bins are omitted, and ECE is reported in percent.
    }
    \label{fig:calibration}
\end{figure*}

\begin{figure}[t]
    \centering
    \includegraphics[width=\columnwidth]
    {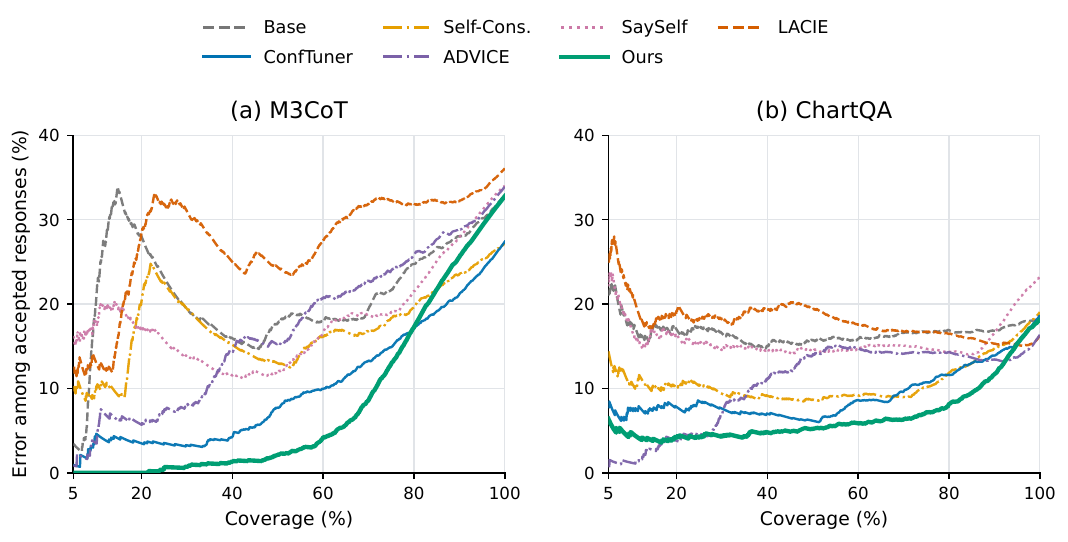}
    \caption{
    Risk--coverage curves on M3CoT and ChartQA.
    Responses are retained in descending confidence order;
    risk is the error rate among retained responses.
    Lower curves indicate better selective prediction.
    The displayed coverage range is 5\%--100\%.
    }
    \label{fig:risk-coverage}
\end{figure}

\myparagraph{Calibration and selective prediction.}
Figures~\ref{fig:calibration} and~\ref{fig:risk-coverage}
visualize two complementary advantages of our confidence estimator.
Our method achieves the lowest ECE among the compared methods on
both datasets, supporting improved agreement between predicted
confidence and observed correctness. The risk--coverage curves
further demonstrate the practical value of these scores for
selecting reliable answers. 


\begin{table}[t]
\centering
\caption{
Performance comparison using Qwen3-VL-8B-Instruct as the foundation model, with training on M3CoT and evaluation on
M3CoT, ChartQA, and MMMU.
Best results are bold, and second-best results are underlined.
}
\scriptsize
\setlength{\tabcolsep}{2.5pt}
\begin{tabular}{llrrrrrrr}
\toprule
\makecell[l]{\textbf{Datasets}}
& \textbf{Methods}
& \textbf{ECE}
& \textbf{NLL}
& \textbf{Brier}
& \textbf{AURC}
& \textbf{EAURC}
& \textbf{AUROC}
& \textbf{ACC} \\
\midrule
\multirow{7}{*}{M3CoT}
& Base & 26.835 & 2.295 & 0.270 & 2.072 & 1.464 & 0.721 & 0.672 \\
& Self-Cons. & 17.969 & 2.163 & 0.204 & 1.744 & 1.335 & 0.694 & \textbf{0.728} \\
& SaySelf & 17.636 & 2.644 & 0.184 & 1.884 & 1.225 & 0.756 & 0.660 \\
& LACIE & 32.974 & 4.549 & 0.328 & 2.712 & 1.964 & 0.573 & 0.639 \\
& ConfTuner & \underline{7.297} & 0.642 & \underline{0.142} & \underline{0.988} & \underline{0.573} & \underline{0.839} & \underline{0.697} \\
& ADVICE & 13.976 & \underline{0.585} & 0.199 & 1.626 & 0.973 & 0.754 & 0.661 \\
& Ours & \textbf{1.316} & \textbf{0.241} & \textbf{0.071} & \textbf{0.757} & \textbf{0.148} & \textbf{0.959} & 0.672 \\
\midrule
\multirow{7}{*}{ChartQA}
& Base & 17.498 & 2.558 & 0.178 & 1.671 & 1.497 & 0.537 & 0.819 \\
& Self-Cons. & 8.838 & 1.228 & 0.132 & 1.082 & 0.888 & 0.751 & 0.809 \\
& SaySelf & 14.140 & 2.029 & 0.150 & 1.594 & 1.300 & 0.718 & 0.768 \\
& LACIE & 15.227 & 2.396 & 0.153 & 1.832 & 1.692 & 0.540 & \textbf{0.838} \\
& ConfTuner & \underline{8.174} & 0.806 & \underline{0.128} & \underline{0.935} & \underline{0.750} & \underline{0.782} & 0.814 \\
& ADVICE & 25.957 & \underline{0.560} & 0.188 & 1.012 & 0.870 & 0.691 & \underline{0.836} \\
& Ours & \textbf{6.532} & \textbf{0.364} & \textbf{0.105} & \textbf{0.677} & \textbf{0.503} & \textbf{0.835} & 0.819 \\
\midrule
\multirow{7}{*}{MMMU}
& Base & 30.256 & 2.582 & 0.309 & 2.641 & 1.778 & 0.697 & \underline{0.614} \\
& Self-Cons. & 19.993 & 2.156 & 0.215 & \underline{2.042} & 1.449 & 0.736 & \textbf{0.676} \\
& SaySelf & 25.700 & 3.958 & 0.260 & 3.652 & 2.071 & 0.748 & 0.493 \\
& LACIE & 37.416 & 5.272 & 0.373 & 3.693 & 2.753 & 0.562 & 0.578 \\
& ConfTuner & 14.983 & 0.960 & \underline{0.211} & 2.148 & \underline{1.257} & \underline{0.773} & 0.609 \\
& ADVICE & \textbf{7.997} & \underline{0.633} & 0.222 & 2.582 & 1.725 & 0.658 & 0.577 \\
& Ours & \underline{9.567} & \textbf{0.562} & \textbf{0.183} & \textbf{1.762} & \textbf{0.899} & \textbf{0.809} & \underline{0.614} \\
\bottomrule
\end{tabular}
\label{tab:vlm}
\end{table}



\myparagraph{Results on language tasks.}
Tables~\ref{tab:llm-qwen} and~\ref{tab:llm-llama} demonstrate consistent
advantages across two LLM backbones and four benchmarks. Our method
achieves the lowest NLL, Brier score, AURC, and EAURC in all eight
model--dataset settings, together with the highest AUROC in seven
settings and a tied best result in the remaining setting.
On HotpotQA, it improves AUROC over the strongest competing baseline
from 0.766 to 0.910 for Qwen3-8B and from 0.814 to 0.907 for
Llama-3.1-8B-Instruct, while also achieving the lowest ECE.
These results demonstrate a joint advantage in assigning accurate
confidence values and distinguishing correct from incorrect answers,
rather than an improvement confined to a single evaluation criterion.

Importantly, this advantage extends beyond the training dataset.
Despite training only on HotpotQA, our method maintains its leading
NLL, Brier score, selective prediction performance, and AUROC on
GSM8K, StrategyQA, and TriviaQA across both backbones. This consistency
supports the transferability of the learned confidence assessment
across mathematical reasoning, commonsense reasoning, and factual
question answering. Moreover, our estimator obtains these gains from the original query and a single generated response, while preserving the generator's answers and accuracy.
This makes it applicable to answers that have already been produced:
confidence assessment can be improved independently, without retraining
the generator or collecting additional answer samples at inference.

\begin{table}[t]
\centering
\caption{
Performance comparison using Qwen3-8B as the foundation model, trained on HotpotQA and evaluated on HotpotQA, GSM8K, StrategyQA, and TriviaQA.
}
\scriptsize
\setlength{\tabcolsep}{2pt}
\begin{tabular}{llrrrrrrr}
\toprule
\makecell[l]{\textbf{Datasets}}
& \textbf{Methods}
& \textbf{ECE}
& \textbf{NLL}
& \textbf{Brier}
& \textbf{AURC}
& \textbf{EAURC}
& \textbf{AUROC}
& \textbf{ACC} \\
\midrule
\multirow{7}{*}{HotpotQA}
& Base & \underline{18.829} & 2.722 & \underline{0.188} & 1.730 & 1.529 & 0.565 & 0.806 \\
& Self-Cons. & 30.148 & 3.969 & 0.305 & 3.076 & 2.238 & 0.631 & 0.620 \\
& SaySelf & 30.782 & 2.579 & 0.308 & 5.499 & 2.353 & 0.702 & 0.321 \\
& LACIE & 64.329 & 10.346 & 0.643 & 1.345 & 1.258 & 0.484 & \textbf{0.871} \\
& ConfTuner & 30.673 & \underline{0.821} & 0.277 & 5.291 & 1.658 & \underline{0.766} & 0.280 \\
& ADVICE & 32.622 & 0.945 & 0.231 & \underline{1.013} & \underline{0.903} & 0.632 & \underline{0.855} \\
& Ours & \textbf{3.465} & \textbf{0.276} & \textbf{0.076} & \textbf{0.457} & \textbf{0.256} & \textbf{0.910} & 0.806 \\
\midrule
\multirow{7}{*}{GSM8K}
& Base & \underline{8.025} & 1.265 & \underline{0.080} & 0.706 & 0.673 & 0.514 & \textbf{0.920} \\
& Self-Cons. & \textbf{3.838} & 0.618 & 0.084 & \underline{0.425} & \underline{0.337} & \underline{0.822} & \underline{0.870} \\
& SaySelf & 10.402 & 1.409 & 0.123 & 0.782 & 0.686 & 0.654 & 0.865 \\
& LACIE & 17.598 & 2.762 & 0.175 & 1.551 & 1.371 & 0.536 & 0.817 \\
& ConfTuner & 11.160 & 2.403 & 0.124 & 1.031 & 0.912 & 0.672 & 0.850 \\
& ADVICE & 13.768 & \underline{0.454} & 0.144 & 0.731 & 0.538 & 0.776 & 0.810 \\
& Ours & 8.573 & \textbf{0.217} & \textbf{0.059} & \textbf{0.139} & \textbf{0.106} & \textbf{0.909} & \textbf{0.920} \\
\midrule
\multirow{7}{*}{StrategyQA}
& Base & 31.727 & 3.987 & 0.321 & 3.004 & 2.378 & 0.536 & 0.668 \\
& Self-Cons. & 28.215 & 4.138 & 0.302 & 3.028 & 2.348 & 0.576 & 0.655 \\
& SaySelf & 15.933 & 1.769 & 0.235 & 2.413 & 1.913 & 0.611 & \textbf{0.701} \\
& LACIE & 38.705 & 5.072 & 0.391 & 2.674 & 2.070 & 0.576 & 0.673 \\
& ConfTuner & 17.307 & 0.836 & 0.244 & 2.546 & 1.845 & \underline{0.645} & 0.650 \\
& ADVICE & \underline{13.523} & \underline{0.738} & \underline{0.225} & \underline{2.305} & \underline{1.756} & 0.618 & \underline{0.687} \\
& Ours & \textbf{6.352} & \textbf{0.557} & \textbf{0.187} & \textbf{1.616} & \textbf{0.990} & \textbf{0.765} & 0.668 \\
\midrule
\multirow{7}{*}{TriviaQA}
& Base & 42.907 & 6.585 & 0.427 & 4.240 & 3.130 & 0.530 & 0.567 \\
& Self-Cons. & 27.223 & 2.474 & 0.258 & 2.939 & 1.604 & \underline{0.776} & 0.530 \\
& SaySelf & 18.355 & 1.987 & 0.229 & 2.567 & 1.615 & 0.744 & \underline{0.597} \\
& LACIE & 45.102 & 7.033 & 0.449 & 4.284 & 3.113 & 0.533 & 0.557 \\
& ConfTuner & 26.744 & 1.558 & 0.271 & 2.565 & 1.684 & 0.745 & \textbf{0.611} \\
& ADVICE & \textbf{11.099} & \underline{0.650} & \underline{0.212} & \underline{2.375} & \underline{1.416} & 0.746 & 0.595 \\
& Ours & \underline{11.639} & \textbf{0.550} & \textbf{0.185} & \textbf{1.933} & \textbf{0.823} & \textbf{0.836} & 0.567 \\
\bottomrule
\end{tabular}
\label{tab:llm-qwen}
\end{table}

\begin{table}[t]
\centering
\caption{
Performance comparison using Llama-3.1-8B-Instruct as the foundation model, trained on HotpotQA and evaluated on HotpotQA, GSM8K, StrategyQA, and TriviaQA.
Best results are bold, and second-best results are underlined.
}
\scriptsize
\setlength{\tabcolsep}{2pt}
\begin{tabular}{llrrrrrrr}
\toprule
\makecell[l]{\textbf{Datasets}}
& \textbf{Methods}
& \textbf{ECE}
& \textbf{NLL}
& \textbf{Brier}
& \textbf{AURC}
& \textbf{EAURC}
& \textbf{AUROC}
& \textbf{ACC} \\
\midrule
\multirow{7}{*}{HotpotQA}
& Base & 18.373 & 1.851 & 0.201 & 1.715 & 1.412 & 0.654 & 0.764 \\
& Self-Cons. & 25.911 & 2.662 & 0.285 & 3.283 & 2.023 & 0.696 & 0.542 \\
& SaySelf & \underline{5.978} & \underline{0.586} & \underline{0.171} & 3.818 & 1.353 & \underline{0.814} & 0.386 \\
& LACIE & 10.903 & 1.237 & 0.178 & \underline{1.093} & \underline{0.833} & 0.743 & \underline{0.780} \\
& ConfTuner & 20.968 & 0.738 & 0.252 & 3.756 & 1.561 & 0.731 & 0.376 \\
& ADVICE & 66.656 & 9.892 & 0.655 & 1.170 & 1.095 & 0.472 & \textbf{0.877} \\
& Ours & \textbf{4.038} & \textbf{0.308} & \textbf{0.090} & \textbf{0.625} & \textbf{0.322} & \textbf{0.907} & 0.764 \\
\midrule
\multirow{7}{*}{GSM8K}
& Base & 16.505 & 1.659 & 0.170 & 1.166 & 0.972 & 0.695 & \underline{0.810} \\
& Self-Cons. & \textbf{4.917} & \underline{0.592} & \underline{0.143} & \underline{0.947} & \underline{0.644} & \underline{0.799} & 0.764 \\
& SaySelf & \underline{5.770} & 0.663 & 0.150 & 1.406 & 1.186 & 0.660 & 0.798 \\
& LACIE & 23.691 & 1.341 & 0.285 & 3.152 & 2.564 & 0.509 & 0.677 \\
& ConfTuner & 18.370 & 2.671 & 0.184 & 1.715 & 1.499 & 0.588 & 0.791 \\
& ADVICE & 43.134 & 6.432 & 0.452 & 1.262 & 1.119 & 0.565 & \textbf{0.823} \\
& Ours & 12.954 & \textbf{0.434} & \textbf{0.140} & \textbf{0.668} & \textbf{0.474} & \textbf{0.824} & \underline{0.810} \\
\midrule
\multirow{7}{*}{StrategyQA}
& Base & 15.352 & \underline{0.764} & 0.221 & 2.157 & 1.698 & 0.605 & \textbf{0.713} \\
& Self-Cons. & 15.197 & 1.831 & 0.199 & \underline{1.774} & \underline{1.290} & \textbf{0.720} & \underline{0.706} \\
& SaySelf & 17.467 & 1.165 & 0.248 & 2.832 & 2.107 & 0.620 & 0.644 \\
& LACIE & 17.616 & 0.948 & 0.247 & 2.486 & 1.749 & 0.660 & 0.641 \\
& ConfTuner & \textbf{6.914} & 0.779 & \underline{0.197} & 2.235 & 1.694 & \underline{0.676} & 0.688 \\
& ADVICE & 30.968 & 3.759 & 0.352 & 2.153 & 1.678 & 0.597 & 0.702 \\
& Ours & \underline{7.739} & \textbf{0.563} & \textbf{0.189} & \textbf{1.598} & \textbf{1.139} & \textbf{0.720} & \textbf{0.713} \\
\midrule
\multirow{7}{*}{TriviaQA}
& Base & 19.465 & 0.856 & 0.227 & 1.742 & 1.212 & 0.736 & 0.692 \\
& Self-Cons. & 8.592 & 0.986 & 0.156 & 1.491 & 0.859 & \underline{0.829} & 0.666 \\
& SaySelf & 8.225 & 0.504 & 0.148 & \underline{1.085} & \underline{0.724} & 0.811 & 0.744 \\
& LACIE & 23.789 & 1.334 & 0.270 & 2.532 & 1.833 & 0.627 & 0.632 \\
& ConfTuner & \underline{4.602} & \underline{0.462} & \underline{0.145} & 1.185 & 0.967 & 0.699 & \textbf{0.757} \\
& ADVICE & 40.995 & 6.408 & 0.452 & 1.571 & 1.317 & 0.565 & \underline{0.746} \\
& Ours & \textbf{4.535} & \textbf{0.407} & \textbf{0.129} & \textbf{1.001} & \textbf{0.470} & \textbf{0.877} & 0.692 \\
\bottomrule
\end{tabular}
\label{tab:llm-llama}
\end{table}


\subsection{Ablation Studies}
\label{sec:ablation_studies}

We examine whether confidence learning benefits from combining individual answer correctness with predictive reliability. Correctness uses only the binary correctness label of the generated answer, while Reliability uses only the estimated probability that the model answers the question correctly. We additionally vary the correctness weight $\alpha$ between 0.3 and 0.7 and compare these settings with our equal-weight combination.

Table~\ref{tab:surrogate-ablation} shows that adding predictive reliability primarily improves the numerical quality of confidence estimates. Compared with correctness-only supervision, our surrogate reduces ECE, NLL, and Brier score on both datasets while preserving AUROC on M3CoT and modestly improving it on ChartQA. These results indicate that adding predictive reliability improves both confidence calibration and selective prediction. Reliability-only supervision yields lower AUROC and worse selective prediction than the combined target on both datasets, supporting the inclusion of information about the particular generated answer. Varying $\alpha$ reveals a trade-off: emphasizing reliability achieves the lowest M3CoT ECE, but weakens discrimination and selective prediction. Equal weighting provides the strongest overall balance among the tested settings.

\begin{table}[t]
\centering
\caption{
Ablation study of the confidence surrogate using Qwen3-VL-8B-Instruct as the foundation model, trained on M3CoT and evaluated on M3CoT and ChartQA.
}
\scriptsize
\setlength{\tabcolsep}{2.5pt}
\begin{tabular}{llrrrrrrr}
\toprule
\makecell[l]{\textbf{Datasets}}
& \textbf{Methods}
& \textbf{ECE}
& \textbf{NLL}
& \textbf{Brier}
& \textbf{AURC}
& \textbf{EAURC}
& \textbf{AUROC}
& \textbf{ACC} \\
\midrule
\multirow{7}{*}{M3CoT}
& Correctness & 4.638 & 0.264 & \underline{0.073} & 0.773 & 0.164 & \textbf{0.959} & \textbf{0.672} \\
& $\alpha=0.7$ & 1.367 & \underline{0.250} & 0.075 & \underline{0.768} & \underline{0.160} & \underline{0.956} & \textbf{0.672} \\
& $\alpha=0.3$ & \textbf{1.258} & 0.255 & 0.075 & 0.778 & 0.169 & 0.954 & \textbf{0.672} \\
& Reliability & 1.667 & 0.272 & 0.082 & 0.798 & 0.190 & 0.948 & \textbf{0.672} \\
& Head only & 3.086 & 0.410 & 0.132 & 1.111 & 0.503 & 0.876 & \textbf{0.672} \\
& Verbalized & 8.954 & 0.797 & 0.101 & 1.019 & 0.411 & 0.924 & \textbf{0.672} \\
& Ours & \underline{1.316} & \textbf{0.241} & \textbf{0.071} & \textbf{0.757} & \textbf{0.148} & \textbf{0.959} & \textbf{0.672} \\
\midrule
\multirow{7}{*}{ChartQA}
& Correctness & 10.103 & 0.466 & 0.121 & 0.693 & 0.519 & 0.826 & \textbf{0.819} \\
& $\alpha=0.7$ & 7.932 & 0.395 & 0.112 & \underline{0.692} & \underline{0.518} & \underline{0.832} & \textbf{0.819} \\
& $\alpha=0.3$ & \underline{6.566} & \underline{0.393} & 0.111 & 0.715 & 0.541 & 0.818 & \textbf{0.819} \\
& Reliability & 7.504 & 0.403 & \underline{0.110} & 0.734 & 0.560 & 0.817 & \textbf{0.819} \\
& Head only & 8.172 & 0.445 & 0.140 & 0.938 & 0.763 & 0.735 & \textbf{0.819} \\
& Verbalized & 12.426 & 1.496 & 0.129 & 1.255 & 1.081 & 0.757 & \textbf{0.819} \\
& Ours & \textbf{6.532} & \textbf{0.364} & \textbf{0.105} & \textbf{0.677} & \textbf{0.503} & \textbf{0.835} & \textbf{0.819} \\
\bottomrule
\end{tabular}
\label{tab:surrogate-ablation}
\end{table}

\myparagraph{Confidence-estimator design.}
We examine two design choices in the confidence estimator: adapting the backbone for confidence estimation and predicting confidence directly as a scalar. We compare our full estimator with two variants. Head only freezes the pretrained backbone and trains only the projection and scalar confidence head, retaining the MSE objective. Verbalized trains the backbone through LoRA to generate numerical confidence as text using token cross-entropy, without a scalar head. Both variants use the same question–response pairs and surrogate target values as our full estimator.

As shown in Table~\ref{tab:surrogate-ablation}, head-only training provides useful confidence estimates, indicating that pretrained representations already contain information relevant to answer correctness. The improvement from backbone adaptation suggests that learning representations specifically for confidence estimation helps the estimator use this information more effectively. The verbalized variant performs worse despite receiving the same surrogate targets, showing that the benefits of supervision also depend on how confidence is learned and predicted. Its comparison with our estimator supports direct scalar prediction with MSE over numerical-text generation with token cross-entropy. These findings motivate both backbone adaptation and scalar regression.

\myparagraph{Effect of the confidence-estimator backbone.}
We vary the estimator's model family and scale to test whether the
framework depends on matching the answer generator's backbone and
whether smaller estimators remain effective. Keeping the training
pairs, surrogate supervision, and evaluation answers fixed isolates
the effect of the estimator backbone.
Supplementary Section D shows that effective
confidence assessment is retained across different model families
and scales, supporting flexible selection of the confidence estimator
independently of the answer generator.

\section{Conclusion}
We introduced a confidence-learning framework for LLMs and VLMs
that combines realized response correctness with query-level predictive
reliability through a graded surrogate. An independent response-conditioned
estimator learns this target and assesses existing answers without
modifying the generator or requiring additional answer samples at
inference. Our theoretical analysis establishes ordering and
mean-preservation properties and connects surrogate prediction error
to selective risk. Experiments demonstrate improvements in calibration,
error discrimination, and selective prediction, including transfer
beyond the training datasets, supporting the value of predictive
reliability as supervision for confidence learning.

\section*{LLM Usage Statement}
We use generative AI tools (ChatGPT) to draft and edit this paper, including the LaTeX of its tables and the source code of its figures.
We also use them when debugging the research code and managing the experiments.
For the related work, these tools help us find references, and we manually checked that every citation is correct.

{
    \small
    \bibliographystyle{ieeenat_fullname}
    \bibliography{main}
}

\clearpage
\setcounter{page}{1}
\maketitlesupplementary
\appendix

\setcounter{tocdepth}{2}
\tableofcontents

\section{Properties of the Confidence Surrogate}
\label{sec:appendix_surrogate_properties}

\subsection{Notation and sampling procedure}
\label{sec:appendix_surrogate_setup}


Fix a response generator and its stochastic decoding configuration.
Let $X$ denote a query, including the image for vision-language tasks.
Conditional on $X=x$, sample $M$ responses independently from the
generator. Write $Y_m$ for response $m$ and $C_m\in\{0,1\}$ for its
reference-based correctness. Define the population predictive reliability
and its empirical estimate as
\begin{equation}
\begin{aligned}
r(x)&=\mathbb{E}[C_m\mid X=x],\\
\widehat r(x)&=\frac{1}{M}\sum_{m=1}^{M}C_m.
\end{aligned}
\label{eq:appendix_reliability_definition}
\end{equation}
The empirical rate characterizes the generator's success across
responses to the query under the fixed decoding configuration,
while $C_m$ records the correctness of the particular response $Y_m$.
For a realized pool, the surrogate for response $Y_m$ is
\begin{equation}
S_m=\alpha C_m+(1-\alpha)\widehat r(X).
\label{eq:appendix_surrogate_definition}
\end{equation}
For each query $x$, we sample one pool of $M$ responses and
use its empirical reliability estimate $\widehat r(x)$
for all surrogate targets in that pool.



\subsection{Correctness ordering}
\label{sec:appendix_correctness_ordering}

For correct and incorrect responses to the same query in a fixed
pool, the shared reliability term cancels, giving
\begin{equation}
s_\alpha(x,y^+)-s_\alpha(x,y^-)=\alpha.
\label{eq:appendix_within_query_ordering}
\end{equation}
Thus, within-query targets strictly distinguish the two outcomes
whenever $\alpha>0$.

Across queries, the general bound $2\alpha-1$ stated in the main
paper remains valid. Our implementation admits a stronger bound
because the training response is selected from the same $M$
responses used to compute empirical reliability. A correct
response implies $\widehat r(x)\geq1/M$, while an incorrect
response implies $\widehat r(x')\leq(M-1)/M$. Therefore,
\begin{equation}
\begin{aligned}
s_\alpha(x,y^+)
&\geq\alpha+\frac{1-\alpha}{M},\\
s_\alpha(x',y^-)
&\leq(1-\alpha)\frac{M-1}{M}.
\end{aligned}
\end{equation}
Consequently,
\begin{equation}
s_\alpha(x,y^+)-s_\alpha(x',y^-)
\geq2\alpha-1+\frac{2(1-\alpha)}{M}>0
\label{eq:appendix_across_query_ordering}
\end{equation}
for $\alpha\in[\frac12,1]$ and finite $M$.
For our setting $M=10$ and $\alpha=\frac12$, correct responses
receive targets of at least $0.55$, while incorrect responses
receive targets of at most $0.45$. Thus, their surrogate targets
are strictly separated, including across different queries.
This property concerns the constructed targets and does not
require training on all $M$ responses.

\subsection{Mean preservation and deviations}
\label{sec:appendix_reliability_preservation}

For any realized response pool, averaging its surrogate targets yields
\begin{equation}
\begin{aligned}
\frac{1}{M}\sum_{m=1}^{M}S_m
&=\frac{\alpha}{M}\sum_{m=1}^{M}C_m
 +(1-\alpha)\widehat r(X)\\
&=\alpha\widehat r(X)+(1-\alpha)\widehat r(X)\\
&=\widehat r(X).
\end{aligned}
\label{eq:appendix_mean_preservation}
\end{equation}
The surrogate therefore changes the targets assigned to individual
responses while preserving the pool's empirical predictive reliability.
Subtracting the shared reference from each target gives
\begin{equation}
S_m-\widehat r(X)=\alpha\bigl[C_m-\widehat r(X)\bigr].
\label{eq:method_surrogate_deviation}
\end{equation}
Hence, $\alpha$ controls the deviation of each constructed target from
the reliability reference. Both identities are algebraic properties of
the constructed targets. They hold even if only one response per query
is subsequently used for confidence training, and impose no exact
averaging constraint on learned predictions.

\subsection{Population target of MSE training}
\label{sec:appendix_population_target}

To characterize the objective, let $J$ be a selected response index,
chosen independently of the sampled response pool conditional on $X$.
Set $Y=Y_J$, $C=C_J$, and
$S=\alpha C+(1-\alpha)\widehat r(X)$. The selected response remains
included in the pool used to compute $\widehat r(X)$.
Define
\begin{equation}
\eta(x,y)=\mathbb{E}[C\mid X=x,Y=y],
\label{eq:appendix_response_correctness_probability}
\end{equation}
the response-level correctness probability under the specified joint
distribution. If correctness is deterministic given the full query,
response, and fixed reference, then $\eta(x,y)=C(x,y)$ almost surely.

For any measurable predictor $c$ with values in $[0,1]$, squared-loss
regression admits the conditional decomposition
\begin{equation}
\begin{aligned}
&\mathbb{E}\bigl[(c(X,Y)-S)^2\mid X,Y\bigr]\\
&\quad=\bigl(c(X,Y)-\mathbb{E}[S\mid X,Y]\bigr)^2
       +\operatorname{Var}(S\mid X,Y).
\end{aligned}
\label{eq:appendix_mse_decomposition}
\end{equation}
Consequently, its unrestricted population minimizer is
$c^*(x,y)=\mathbb{E}[S\mid X=x,Y=y]$.
Under conditionally independent sampling of the response--correctness
pairs and the outcome-independent index selection above,
\begin{equation}
\mathbb{E}[\widehat r(X)\mid X=x,Y=y]
=\frac{\eta(x,y)+(M-1)r(x)}{M}.
\label{eq:appendix_conditional_reliability}
\end{equation}
Indeed, the selected response contributes the conditional mean
$\eta(x,y)$; each of the other $M-1$ responses contributes $r(x)$.
Substituting into the surrogate yields
\begin{equation}
\begin{aligned}
c^*(x,y)&=\beta_M\eta(x,y)+(1-\beta_M)r(x),\\
\beta_M&=\alpha+\frac{1-\alpha}{M}.
\end{aligned}
\label{eq:appendix_population_optimum}
\end{equation}
For $M=10$ and $\alpha=\frac12$, $\beta_M=0.55$.
The finite-$M$ correction accounts for the selected response contributing
to its own query's empirical reliability estimate. A reliability pool
sampled independently of the selected training response would instead
give the mixture weight $\alpha$.

This characterization identifies the population target as shrinkage of
response-specific correctness toward query reliability. It generally
differs from $\eta(x,y)$, so calibration against correctness does not
follow from the target construction alone. It is an unrestricted
population optimum, rather than a guarantee of convergence or
representational capacity for a particular neural estimator. Practical
benefits are assessed through the experimental comparisons and ablations.
The finite-$M$ formula requires outcome-independent selection; selecting
responses based on their correctness or the realized pool can change it.

\section{Selective-Risk Analysis}
\label{sec:appendix_selective_risk}

\subsection{Definitions and guarantee}

Fix a trained confidence estimator $c_\theta(X,Y)\in[0,1]$ and a joint
evaluation distribution over queries, generated responses, and any
sampling randomness used to construct their surrogate targets.
The surrogate target is used here to define population prediction error;
it is not an input required for inference. Let
\begin{equation}
\begin{aligned}
\varepsilon_\alpha
&=\mathbb{E}\bigl[(c_\theta(X,Y)-S)^2\bigr],\\
S&=\alpha C+(1-\alpha)\widehat r(X).
\end{aligned}
\label{eq:appendix_population_surrogate_error}
\end{equation}
For an acceptance threshold $\tau$, define coverage and selective risk as
\begin{equation}
\begin{aligned}
\phi_\tau&=\Pr\bigl(c_\theta(X,Y)\geq\tau\bigr),\\
R_\tau&=\Pr\bigl(C=0\mid c_\theta(X,Y)\geq\tau\bigr).
\end{aligned}
\label{eq:appendix_selective_definitions}
\end{equation}
All expectations and probabilities refer to the same evaluation
distribution.

\paragraph{Theorem (Surrogate-to-selective-risk bound).}
For $\alpha\in(0,1]$ and any threshold
$\tau\in(1-\alpha,1]$ with $\phi_\tau>0$,
\begin{equation}
R_\tau\leq
\frac{\varepsilon_\alpha}
{\phi_\tau[\tau-(1-\alpha)]^2}.
\label{eq:appendix_selective_risk_bound}
\end{equation}

\paragraph{Proof.}
For an incorrect response, $C=0$, so
$S=(1-\alpha)\widehat r(X)\leq1-\alpha$.
If this response is accepted, then
\begin{equation}
c_\theta(X,Y)-S\geq\tau-(1-\alpha)>0.
\end{equation}
Thus, every accepted incorrect response incurs squared surrogate error
at least $[\tau-(1-\alpha)]^2$. Taking expectations gives
\begin{equation}
\begin{aligned}
\varepsilon_\alpha
&\geq[\tau-(1-\alpha)]^2\\
&\quad\cdot\Pr\bigl(C=0,\ c_\theta(X,Y)\geq\tau\bigr)\\
&=[\tau-(1-\alpha)]^2\phi_\tau R_\tau.
\end{aligned}
\end{equation}
Dividing by the positive factor
$\phi_\tau[\tau-(1-\alpha)]^2$ proves the result.
\hfill$\square$

For $\alpha=\frac12$, the guarantee applies to $\tau>\frac12$.
The argument uses only the upper bound on surrogate targets for
incorrect responses; conditional independence of the response pool is
not required for this bound. The guarantee depends on population error.
Low empirical training loss alone does not establish a small
$\varepsilon_\alpha$, and the bound is informative only when its
right-hand side is sufficiently small.



\subsection{Relation to binary correctness supervision}

Setting $\alpha=1$ recovers the corresponding guarantee for
binary correctness supervision. Although this case has a smaller
multiplicative constant at a common admissible threshold,
the prediction errors are measured against different targets,
and separately trained estimators may have different coverage.
The bounds therefore do not establish a performance ordering.
Our result shows that the proposed surrogate retains
selective-risk control, while the empirical comparison evaluates
its practical benefit over correctness-only supervision.

\section{Implementation Details}
\label{sec:appendix_implementation}

\subsection{Construction of training supervision}

For each training query, we generate a pool of $M=10$ responses
using the corresponding answer-generating model. Predictive
reliability is estimated as the fraction of correct responses
under that model's sampling configuration. We select one response
per query using a reproducible query-specific seed derived from
the query identifier. Its surrogate
target combines its binary correctness with the estimated
reliability using $\alpha=\frac12$. The selected response remains
included in the reliability pool. 

\paragraph{M3CoT with Qwen3-VL-8B-Instruct.}
We generate the training response pools using
Qwen3-VL-8B-Instruct, following its official vision-language
sampling recommendations: temperature $0.7$, top-$p$ $0.8$,
top-$k$ $20$, repetition penalty $1.0$, and presence penalty $1.5$.
We set the maximum response length to $1{,}024$ tokens and the
context limit to $4{,}096$ tokens.
Each prompt contains the question and its supplied image and
requests reasoning under \texttt{Thinking:}, followed by a final
\texttt{Answer:} line.

We determine correctness by extracting the predicted option and
comparing it with the reference option. Extraction prioritizes
the final line-start \texttt{Answer:} field, which expects a
parenthesized option letter. A malformed final \texttt{Answer:}
field is labeled incorrect. When this field is absent, the parser
supports an \texttt{<answer>} tag, a leading option, or an
unambiguous explicit answer statement. Empty, unparseable, and
invalid responses count as incorrect and remain in the reliability
denominator. The same extraction rules are used for M3CoT evaluation.

\paragraph{HotpotQA with Qwen3-8B.}
We construct a separate set of response pools using Qwen3-8B. We use the officially recommended temperature
$0.7$, top-$p$ $0.8$, and top-$k$ $20$, and additionally set the
repetition penalty to $1.0$ and the presence penalty to $1.5$.
The maximum response length is $1{,}024$ tokens, the context limit
is $16{,}384$ tokens.
Each generation prompt contains the HotpotQA question and its
accompanying documents, followed by an instruction to provide
reasoning and a short final answer. The ground-truth answer is
withheld during response generation and supplied to the
correctness judge when constructing surrogate targets.

Binary correctness labels are assigned by a Qwen3-8B judge. This judge receives the original question and context,
the reference answer, and the complete generated response.
It assesses semantic equivalence between the final committed
answer and the reference, accepting equivalent aliases,
paraphrases, and numerical or date expressions. A correct answer
mentioned only in discarded reasoning does not count as correct.
Missing answers, abstentions, materially incomplete answers, and
unresolved contradictory alternatives are labeled incorrect.

\paragraph{HotpotQA with Llama-3.1-8B-Instruct.}
We generate a separate set of response pools using
Llama-3.1-8B-Instruct with temperature $0.6$, top-$p$ $0.9$,
and top-$k$ filtering disabled. The repetition penalty is $1.0$;
the presence and frequency penalties are zero. We use a maximum
response length of $1{,}024$ tokens, a context limit of
$16{,}384$ tokens.
For each HotpotQA training question, we provide
Llama-3.1-8B-Instruct with the question and accompanying
documents and sample $M=10$ responses to construct
its surrogate target.

Training correctness labels are assigned by the Qwen3-8B judge
using the same inputs and semantic-equivalence rubric. Missing answers count as incorrect.

\subsection{Estimator input and confidence readout}

All three estimators receive the original query and the complete
selected response, including reasoning when present. M3CoT inputs
retain the original image; HotpotQA inputs retain the question
and supplied context. The estimator input is organized as
\begin{quote}
\small
\textbf{User:} $\langle\text{query}\rangle$\\
\texttt{Answer:} $\langle\text{generated response}\rangle$\\
\texttt{How confident are you in your previous answer?}\\
\textbf{Assistant:} \texttt{Confidence:}
\end{quote}

\paragraph{M3CoT estimator.}
The estimator backbone is
\texttt{Qwen/Qwen3-VL-8B-Instruct}, with the
\texttt{qwen3\_vl} chat template. It encodes the query, image,
supplied response, confidence question, and assistant prefix.

\paragraph{Qwen3-8B HotpotQA estimator.}
The estimator backbone is \texttt{Qwen/Qwen3-8B}, with the
\texttt{qwen3} chat template and thinking disabled.
Its input contains only text.

\paragraph{Llama-3.1-8B-Instruct HotpotQA estimator.}
The estimator backbone is
\texttt{meta-llama/Llama-3.1-8B-Instruct}, with the
\texttt{llama3} chat template. Its input contains only text.

For each backbone, we read the final decoder-layer representation
at the last non-padding token of the serialized confidence input.
Adaptive layer weighting is disabled. Let $d_h$ denote the
backbone-specific hidden dimension. All three estimators use
the same confidence-head structure:
\begin{equation}
\begin{aligned}
\mathbf u(x,y)
&=\operatorname{GELU}\!\left(
W_p\operatorname{LN}(\mathbf h(x,y))\right),\\
c_\theta(x,y)
&=\sigma\!\left(\mathbf w_o^\top\mathbf u(x,y)+b_o\right),
\end{aligned}
\label{eq:appendix_confidence_readout}
\end{equation}
where $W_p\in\mathbb R^{256\times d_h}$ is a bias-free projection,
$\operatorname{LN}$ is layer normalization, and the scalar output
layer includes a bias. The confidence head has no added dropout. Confidence is returned directly as a scalar.

\subsection{Training configuration}

For each estimator, language-model LoRA adapters and the complete
confidence head are optimized jointly using mean squared error:
\begin{equation}
\mathcal L_{\mathrm{sur}}(\theta)
=\frac{1}{N}\sum_{i=1}^{N}
\bigl[c_\theta(x_i,y_i)-s_\alpha(x_i,y_i)\bigr]^2.
\label{eq:appendix_training_objective}
\end{equation}
The original answer-generating model remains fixed;
confidence training updates the estimator's adapters and head.
We use AdamW with weight decay $0.1$ and a cosine learning-rate
schedule with a warmup ratio of $0.05$.
LoRA dropout is set to $0.05$.

\paragraph{Qwen3-VL-8B-Instruct on M3CoT.}
We use LoRA rank $8$, alpha $32$, learning rate $10^{-4}$,
and warmup ratio $0.05$. LoRA targets the language-model linear
modules; the vision encoder and multimodal aligner are frozen.
The full confidence head is saved with the adapters.
Training runs for three epochs with two GPU workers, four
examples per GPU, and four gradient-accumulation steps, giving
an effective batch size of $32$. The maximum training length is
$4{,}096$ tokens, and attention uses FlashAttention.

\paragraph{Qwen3-8B on HotpotQA.}
We use LoRA rank $8$, alpha $32$, learning rate $10^{-4}$,
and warmup ratio $0.05$. LoRA targets all language-model linear
modules.
The reported run uses two epochs,
two examples per GPU, $16$ gradient-accumulation
steps, and two workers, giving an effective
batch size of 64.
The maximum training length is $16{,}384$ tokens, and attention
uses FlashAttention.

\paragraph{Llama-3.1-8B-Instruct on HotpotQA.}
We use LoRA rank $8$, alpha $32$, learning rate $10^{-4}$,
and warmup ratio $0.05$. LoRA targets all language-model linear
modules.
The reported run uses two epochs,
two examples per GPU, $16$ gradient-accumulation
steps, and two workers, giving an effective
batch size of 64.
The maximum training length is $16{,}384$ tokens, and attention
uses FlashAttention.



\paragraph{Software and hardware.}
We use ms-swift $4.5.3$ and train each estimator on two
NVIDIA H100 GPUs, each with $80$~GB of HBM3 memory.

\subsection{Generation and evaluation protocols}

\paragraph{M3CoT.}
Evaluation uses one saved response per query generated by
Qwen3-VL-8B-Instruct with temperature $0.7$, top-$p$ $0.8$,
top-$k$ $20$, repetition penalty $1.0$, presence penalty $1.5$, maximum response length $1{,}024$,
and context limit $4{,}096$. The generation engine is vLLM
through ms-swift, with bfloat16 precision. The estimator scores the saved
response, and correctness is determined by the option-extraction
procedure described above.

\paragraph{HotpotQA with Qwen3-8B.}
The labeled HotpotQA distractor validation split supplies
in-domain evaluation queries. We generate one response per
query using Qwen3-8B in non-thinking mode with temperature
$0.7$, top-$p$ $0.8$, top-$k$ $20$, repetition penalty $1.0$,
presence penalty $1.5$, maximum response
length $1{,}024$, and context limit $16{,}384$.
Evaluation correctness is assigned by \texttt{gpt-6-luna}. Its rubric matches the training
judge's semantic-equivalence criteria. 

\paragraph{HotpotQA with Llama-3.1-8B-Instruct.}
We generate one response per validation query using
Llama-3.1-8B-Instruct with temperature $0.6$, top-$p$ $0.9$,
top-$k$ filtering disabled, repetition penalty $1.0$,
zero presence and frequency penalties, maximum response length $1{,}024$, and context limit $16{,}384$.
Evaluation correctness is assigned by
\texttt{gpt-6-luna}, using the same
semantic-equivalence rubric used for Qwen3 evaluation.

\paragraph{ChartQA evaluation.}
ChartQA is used only for evaluation. 
It retains all $2{,}500$ questions, comprising $1{,}250$
human-written and $1{,}250$ augmented questions.
Each prompt contains the image, question, and an instruction
to return \texttt{Thinking:} reasoning followed by an
\texttt{Answer:} line.
Qwen3-VL-8B-Instruct generates one response per query using
temperature $0.7$, top-$p$ $0.8$, top-$k$ $20$, repetition
penalty $1.0$, presence penalty $1.5$.
For ChartQA, we use a maximum response length of $8{,}192$
tokens and a context limit of $32{,}768$ tokens.
The minimum and maximum image pixel counts are set to
$1{,}003{,}520$ and $4{,}014{,}080$, respectively. 


\paragraph{MMMU evaluation.}
We evaluate on the $900$ MMMU validation questions using
Qwen3-VL's released evaluation prompt. We generate one response
per question with Qwen3-VL-8B-Instruct using temperature $0.7$,
top-$p$ $0.8$, top-$k$ $20$, repetition penalty $1.0$,
presence penalty $1.5$.
The maximum response length is $32{,}768$ tokens, and the
context limit is $128{,}000$ tokens. The minimum and maximum
image pixel counts are $1{,}003{,}520$ and $4{,}014{,}080$,
respectively. We evaluate the M3CoT-trained confidence
estimator without further training on MMMU.
Correctness evaluation follows Qwen3-VL's released
rule-based option extraction, with GPT-6 Luna resolving
responses that cannot be matched by the extraction rules.


\paragraph{Text transfer evaluation.}
We evaluate the HotpotQA-trained confidence estimators on
GSM8K, StrategyQA, and TriviaQA without further training.
We use the GSM8K test split ($1{,}319$ questions), the
TriviaQA validation split
($11{,}313$ questions), and the publicly labeled StrategyQA
training set ($2{,}290$ questions).

Each prompt requests reasoning under \texttt{Thinking:},
followed by a final answer under \texttt{Answer:}.
The requested final answer is numerical for GSM8K,
yes or no for StrategyQA, and a short answer span for TriviaQA.
The confidence estimator receives the original query and
the complete generated response, including its reasoning
and final answer.

For each query, we generate one response using the original
Qwen3-8B or Llama-3.1-8B-Instruct answer generator and score it
with the corresponding HotpotQA-trained confidence estimator.
For Qwen3-8B, we use temperature $0.7$, top-$p$ $0.8$,
top-$k$ $20$, repetition penalty $1.0$, and presence
penalty $1.5$. For Llama-3.1-8B-Instruct, we use temperature
$0.6$, top-$p$ $0.9$, disable top-$k$ filtering, and set the
repetition penalty to $1.0$ and the presence and frequency
penalties to zero. Both generators use a maximum response
length of $1{,}024$ tokens and a context limit of
$16{,}384$ tokens.

Correctness is assigned by \texttt{gpt-6-luna} using the same
semantic-equivalence rubric as HotpotQA evaluation. For TriviaQA, the reference supplied to the judge
includes the canonical answer and accepted aliases.
Missing responses count as incorrect. These evaluations
use a shared semantic-correctness protocol across datasets.

\subsection{Inference and confidence metrics}

For all three backbones, the estimator scores the original
query and one supplied response in a forward pass.
Reference answers and repeated sampling are required for
surrogate construction rather than confidence scoring.
The estimator does not require the answer generator's
parameters, hidden states, or token probabilities.

Confidence metrics compare scalar predictions with binary
response correctness rather than surrogate targets.
ECE uses ten equal-width bins. Confidence is clipped to
$[10^{-7},1-10^{-7}]$ only for NLL; the Brier score uses the
original confidence values. AURC averages the cumulative
error rate over retained prefixes ordered by decreasing
confidence, with stable ordering for ties. EAURC subtracts
the continuous oracle area $e+(1-e)\log(1-e)$, where $e$ is
the error rate, taking the limiting value one at $e=1$.
AUROC treats correct responses as the positive class.

\section{Effect of the confidence-estimator backbone.}

We develop an independent confidence estimator that assesses existing model-generated answers from the original query and a single generated response. To examine whether the framework depends on a particular backbone,
we train confidence estimators from three model families at different
scales: the 8B, 4B, and 2B instruction-tuned variants of Qwen3-VL;
Gemma-3-4B-IT~\cite{team2025gemma}; and the 8B and 2B variants of
InternVL3.5~\cite{wang2025internvl3}. All estimators use the same M3CoT training pairs generated by Qwen3-VL-8B-Instruct, the same surrogate supervision, and the same saved evaluation answers.

Table~\ref{tab:backbone-ablation} shows that the framework remains effective when the confidence estimator uses a different backbone from the answer generator. InternVL3.5-8B closely matches Qwen3-VL-8B-Instruct on M3CoT and achieves higher AUROC on ChartQA, indicating that effective assessment does not require matching their architectures. Smaller backbones also achieve competitive results. These results support flexibility in selecting the confidence-estimator backbone.

\begin{table}[t]
\centering
\caption{
Ablation study of the confidence estimator architecture, trained on M3CoT and evaluated on M3CoT and ChartQA.
}
\vspace{-.1in}
\scriptsize
\setlength{\tabcolsep}{1.5pt}
\begin{tabular}{llrrrrrrr}
\toprule
\makecell[l]{\textbf{Datasets}}
& \textbf{Methods}
& \textbf{ECE}
& \textbf{NLL}
& \textbf{Brier}
& \textbf{AURC}
& \textbf{EAURC}
& \textbf{AUROC}
& \textbf{ACC} \\
\midrule
\multirow{6}{*}{M3CoT}
& Qwen3-VL-8B-Ins & 1.316 & \textbf{0.241} & \textbf{0.071} & \textbf{0.757} & \textbf{0.148} & \textbf{0.959} & \textbf{0.672} \\
& Qwen3-VL-4B-Ins & 1.836 & 0.277 & 0.083 & 0.800 & 0.192 & 0.948 & \textbf{0.672} \\
& Qwen3-VL-2B-Ins & \textbf{1.149} & 0.288 & 0.087 & 0.835 & 0.226 & 0.941 & \textbf{0.672} \\
& Gemma-3-4B-IT & 2.547 & 0.310 & 0.094 & 0.876 & 0.267 & 0.932 & \textbf{0.672} \\
& InternVL3.5-8B & \underline{1.227} & \underline{0.257} & \underline{0.077} & \underline{0.775} & \underline{0.167} & \underline{0.954} & \textbf{0.672} \\
& InternVL3.5-2B & 4.004 & 0.343 & 0.105 & 0.896 & 0.288 & 0.921 & \textbf{0.672} \\
\midrule
\multirow{6}{*}{ChartQA}
& Qwen3-VL-8B-Ins & 6.532 & 0.364 & 0.105 & \textbf{0.677} & \textbf{0.503} & \underline{0.835} & \textbf{0.819} \\
& Qwen3-VL-4B-Ins & 6.105 & 0.360 & 0.102 & 0.801 & 0.626 & 0.819 & \textbf{0.819} \\
& Qwen3-VL-2B-Ins & \textbf{3.578} & \textbf{0.338} & \underline{0.098} & 0.731 & 0.556 & 0.825 & \textbf{0.819} \\
& Gemma-3-4B-IT & \underline{4.157} & 0.375 & 0.110 & 0.865 & 0.691 & 0.783 & \textbf{0.819} \\
& InternVL3.5-8B & 6.483 & 0.351 & \textbf{0.097} & \underline{0.683} & \underline{0.509} & \textbf{0.853} & \textbf{0.819} \\
& InternVL3.5-2B & 4.534 & \underline{0.348} & \underline{0.098} & 0.760 & 0.586 & 0.822 & \textbf{0.819} \\
\bottomrule
\end{tabular}
\label{tab:backbone-ablation}
\vspace{-.2in}
\end{table}




\end{document}